\documentclass{article}
\usepackage{amsmath}
\usepackage[utf8]{inputenc}
\usepackage{geometry}
 \usepackage{graphicx}
 \usepackage{titling}

 \usepackage{authblk}

\usepackage{lipsum}  
\usepackage{cmbright}

\begin{document}

\title{Comparing Fairness of Generative Mobility Models}

\author[1]{Daniel Wang, Jack  McFarland, Afra Mashhadi\footnote{Corresponding author: mashhadi@uw.edu}}
\author[2]{Ekin Ugurel}
\affil[1]{Computing and Software Systems, University of Washington, Bothell, USA}
\affil[2]{Department of Civil \& Environmental Engineering, University of Washington, Seattle, USA}

\date{}

\vspace{-0.5cm}

\maketitle
\vspace{-2cm}
\section*{Extended Abstract}

    Crowd flow data serves as a foundation for predictive modeling, aiding in understanding of city structure~\cite{bassolas2019hierarchical}, epidemics~\cite{wesolowski2012quantifying}, and visitation patterns~\cite{schlapfer2021universal}. Extensive mobility studies have applied variety of models in this area, however there  has been \textbf{little work in quantifying the fairness of them in terms of the equity of their performance across geographical space}.  Indeed, as recent evidence from the broader machine learning domain has shown, the systematic discrimination in making decisions against different groups has been shifted from people to autonomous algorithms~\cite{kasy2021fairness}.  In many applications, discrimination may be defined by different protected attributes, such as race, gender, ethnicity, and religion, that directly prevent favorable outcomes for a minority group in societal resource allocation, education equality, employment opportunity, etc~\cite{sattigeri2019fairness}. Similarly, in the context of spatiotemporal data, mobility demand prediction algorithms have been shown to offer higher service quality to neighborhoods with a larger Caucasian population~\cite{brown2018ridehail}. However, in such contexts, only a handful of recent studies exist that examine the fairness of location-based systems~\cite{yan2020fairness}, and to the best of our knowledge, no effort has yet been made in measuring the fairness of generative mobility models and their resulting synthetic traces. In this paper, we seek to measure and compare these models based on fairness performance across multiple datasets.

We do so by introducing metrics for evaluating  both the utility of a generative model and the quality and fairness of their resulting traces, where accuracy is measured in terms of \textbf{Common Part of Commuters (CPC)}. This metric is based on the Sørensen-Dice index and measures the similarity between real flows and generated flows (Equation 1). 
\begin{equation}
CPC=\frac{2\Sigma_{i,j} (min(\hat{y}(i,j),{y}(i,j))}{\Sigma_{i,j}\hat{y}(i,j) + \Sigma_{i,j}{y}(i,j)}
\end{equation}

Where $\hat{y}$ is the generated flow, and $y$ is the real flow between location i and j.  It is worth noting that, $CPC$ is always positive and contained in the closed interval $[0, 1]$ where 1 is a perfect match between the generated flows and the ground truth and 0 is a poor performance with no  overlap between generated flows and real flows. 

Fairness is measured based on   \textbf{Demographic Parity}~\cite{friedler2021possibility} that is the demographic groups should receive similar decisions, inspired by civil rights laws in different countries~\cite{barocas2016big}.
To be specific, group fairness argues that a disadvantaged and advantaged groups (in terms of the sensitive attributes) should receive similar treatments:

\begin{equation}
\begin{aligned}
    P(Y'= 1|S = 0, Y = 1) = P(Y' = 1|S = 1, Y = 1)
\end{aligned}
\label{equ:group fairness}
\end{equation}

That is, the positive outcome of a classifier $Y'$ should occur with the same probability regardless of the sensitive attribute $S$, often referring to a binary property such as being above or below an income threshold. In this study, we  reformulate the above equation to compare the probability distribution of CPCs for a subset of $S$ (i.e., $P(CPC,S)$) and $S'$ (i.e., $P(CPC,S')$), where $S'$ corresponds to the disadvantaged group  and $S$ corresponds to the advantaged group.  For group fairness to be satisfied, we expect the distance between these two probability distributions to be close to zero. That is a generative model would create synthetic traces for richer and poorer areas of the city with the same error ratio. 

To measure the distance between probabilities we use Kullback Leibler (KL) divergence. KL-Divergence distance results in zero if two probability distributions are identical, indicating group fairness. 

\subsection*{Summary of Findings}

We use four mobility models, namely, Gravity~\cite{pappalardo2019scikit}, Radiation~\cite{simini2012universal}, Deep Gravity and Non-linear Gravity both by~\cite{simini2021deep} over  New York Taxi Dataset~\cite{taxinyc2009}. This dataset contains data collected in New York City, which was collected between 2016/4/1 and 2016/6/30 and contains over 35 million records. 
In order to measure group fairness metrics as defined earlier, we need data on socio-economic level of geographical neighborhoods. To this end, we use the Social Vulnerability Index (SVI). SVI was developed by the Centers for Disease Control and Prevention as a metric that captures the resilience of each community at the census tract level. This index combines 15 U.S. census variables grouped into four themes: \textbf{socioeconomic} status,\textbf{ household }characteristics, \textbf{race and ethnicity}, and  \textbf{ housing and transportation},  in order to rank census tracts by their relative vulnerability to hazardous events. Each of these themes has its own percentile ranking where  greater percentile values represent greater \textit{vulnerability}.   We use  group origin-destination pairs that belong to the bottom interquartile range or top-interquartile range of the overall SVI theme and its sub-themes to present $S$ and $S'$ respectively.

\begin{table}[]
\centering
\begin{tabular}{lllll}
\multicolumn{1}{c}{} & \multicolumn{1}{c}{Gravity} & \multicolumn{1}{c}{Radiation} & \multicolumn{1}{c}{Deep gravity} & \multicolumn{1}{c}{Non-Linear Gravity} \\
Mean CPC & (0.35) & (0.24) & (\textit{0.41}) & (0.36) \\  
\hline
\hline
\textbf{Fairness-SVI-Total}            & \textit{0.60 }                       & 0.97                          & 4.10                             & 3.34                                   \\
~~{Fairness-Socioeconomic}        & \textit{0.68}                        & 0.99                          & 4.87                             & 3.51                                   \\
~~{Fairness-Household}            & \textit{0.61 }                       & 1.01                          & 5.94                             & 4.82                                   \\
~~{Fairness-Ethnicity}            & \textit{0.70}                        & 0.92                          & 8.22                             & 7.86                                   \\
~~{Fairness-Transportation}       & \textit{0.65}                        & 0.88                          & 5.43                             & 4.13                                  
\end{tabular}
\caption{Group Fairness for NY Dataset for various models across  SVI indicators. The lowest values (shown in italic)  for SVI-total and its  sub-themes present the fairer model in creating more equitable synthetic traces.   }
\label{tab}
\end{table}

Table~\ref{tab} presents the group fairness of each model computed over SVI and its sub-themes for the NY dataset. The table reports the goodness of generated traces for all areas as mean CPC. As expected and similar to the results reported in~\cite{simini2021deep}, Deep Gravity outperforms other models in terms of CPC. However, regarding group fairness, we can see that the traditional gravity model performs the fairest for all the SVI themes followed by the Radiation model. The deep Gravity model in comparison to other models performs much worse in terms of group fairness, even though it achieves higher average CPC. 
Comparing non-linear gravity and deep gravity we can see that the feature embedding part of the deep gravity model that results in creating better synthetic traces also leads to unwanted biases.  We hypothesize that this may be caused by existing biases in the POI representation of the poorer communities in OSM.

\vspace{-0.5cm}
\section*{Acknowledgments}

This work was supported by generous support from the US National Science Foundation (NSF) award \textbf{IIS-2304213}.

\vspace{-0.5cm}
\footnotesize


\end{document}